# Enhancing Ayurvedic Diagnosis using Multinomial Naïve Bayes and K-modes Clustering: An Investigation into Prakriti Types and Dosha Overlapping

**Pranav Bidve[1], Shalini Mishra[2], and Annapurna J[3].**
Vellore Institute of Technology, Vellore
[1]bidvepranav57@gmail.com, [2]shamishra2002@gmail.com, [3]annapurna.ca3@gmail.com

**ABSTRACT** The identification of 'Prakriti' types for the human body is a long-lost medical practice in finding the harmony between the nature of human beings and their behaviour. There are 3 fundamental 'Prakriti' types of individuals. A person can belong to any 'Dosha'. In the existing models, researchers have made use of SVM, KNN, PCA, Decision Tree, and various other algorithms. The output of these algorithms was quite decent, but it can be enhanced with the help of Multinomial Naïve Bayes and K-modes clustering. Most of the researchers have confined themselves to 3 basic classes. This might not be accurate in the real-world scenario, where overlapping might occur. Considering these, we have classified the 'Doshas' into 7 categories, which includes overlapping of 'Doshas'. These are namely, 'VATT-Dosha', 'PITT-Dosha', 'KAPH-Dosha', 'VATT-PITT-Dosha', 'PITT-KAPH-Dosha', 'KAPH-VATT-Dosha', and 'VATT-PITT-KAPH-Dosha'. The data used contains a balanced set of all individual entries on which preprocessing steps of machine learning have been performed. Chi-Square test for handling categorical data is being used for feature selection. For model fitting, the method used in this approach is K-modes clustering. The empirical results demonstrate a better result while using the MNB classifier. All key findings of this work have achieved 0.90 accuracy, 0.81 precision, 0.91 F-score, and 0.90 recall. The discussion suggests a provident analysis of the seven clusters and predicts their occurrence. The results have been consolidated to improve the Ayurvedic advancements with machine learning.

**INDEX TERMS** Ayurveda, 'Prakriti' types, vatt, pitt, kaph, chi-square test, machine learning

## I. INTRODUCTION

The 'Prakriti' type of an individual can be a portal to understanding the eclectic behaviour of humanity. The study not only involves studying a particular individual based on the body type but also takes into consideration his/her body vitals like normal heart rate, number of hours of sleep, number of steps walked, etc, …The data retrieved from the body then can be aggregated with the 'Prakriti' type to better understand the physical aspect of an individual. Based on the obtained results, we can then go ahead to give medical recommendations for a particular or general diagnosis problem. An automated machine learning model is employed to continuously study and account for nuanced differences to suit a person. The research work focuses on all three types – 'Ekadoshaja', 'Dwandwaja', and 'Tridoshaja'. According to Ayurveda, they have been named and recognized as 'Vatt', 'Pitt', and 'Kaph'. However, in this work, we shall be naming them as 'Vata', 'Pita', and 'Kapha'.

### A. MOTIVATION AND JUSTIFICATION

The purpose of this study is the augmentation of already set up studies pertinent to ayurveda but insufficient because they lack real-world applications. This study aims at recognizing past works, analysing the myopic insights of the already established studies, and suggesting better ways to incorporate Ayurveda in our daily life. In this work, we will be focusing on all these categories, work on their categorization and eventually construct a model that implicitly recognizes the 'Prakriti' types of a person. We aim to improvise the very core of past works by using a dataset that includes an even demographic distribution. This would not limit us to a mere population of a locality having people with consonant lifestyles. To gather a deeper insight into smaller hierarchical clusters of the three fundamental clusters, we fuse 'Dwandawaja' which uses seven classes into our paper in contrast to traditional 'Ekadoshaja' which uses three classes.

### B. LITERATURE SURVEY

Ayurveda is one of the oldest medical sciences, which originated more than five thousand years ago in the Indian subcontinent [1]. It is a boon to healthcare due to its holistic system of medicine that considers the whole person, including their physical, mental, and emotional health. These features help in assessing the health of a person. An important classification needs to be done which is based on multiple features and characteristics of the person. These classifications are called 'Dosha'. 'Dosha' refers to the three fundamental energies or functional principles that govern the human body and mind. According to Ayurvedic philosophy, each person has a unique combination of these 'Doshas', which determine their physical, mental, or emotional characteristics, also their predisposition to certain health conditions. There has been some research that has already addressed this problem and has built a model to make the classification into doshas successful with the help of different algorithms.

Support Vector Machine (SVM) used by Vishu madaan at.al., T. Thanushree at.al., [2] and Md. Gulzaar Hussain at.al.,[3] for classification. SVM offers a faster prediction, better accuracy, and uses less memory when compared to traditional algorithms. Vishnu Madaan used this to predict the Prakriti type of person. He classified them into 7 different classes with the help of this algorithm. Using SVM on our dataset seems feasible because its complexity entirely depends on the number of data points, which is quite less in this case. However, there are a few disadvantages due to which we will not be making use of this algorithm, most significant factor is overlapping, there is a lot of overlap in the target classes, which causes a lot of ambiguity when SVM is used. Hence it will be safe to avoid SVM in our research.

Another model used by [1], W. Cherif at.al.,[4] and T. Thanushree at.al.,[2] is KNN, a traditional algorithm, which calculates the distance between 2 points, and then classifies based on the nearness to the point. This algorithm can be used on a small dataset, as it would become easier to find the distance, and the time complexity would reduce. One disadvantage of KNN is that it is sensitive to scaling, which means that if there are more features in the dataset, that are not affecting the prediction of the mode, then KNN will not be efficient for prediction. Hence, we will not be making use of KNN algorithm in our research.

Principle Component Analysis (PCA) has been used by Periyaswamy Govindaraj at.al.,[6]. In his research, he had performed genome-wide single nucleotide polymorphism analysis on 262 men, and with the help of this data, he classified them into 'Prakriti. He made use of PCA, In which the important or necessary features can be extracted, to reduce the dimensions, which would not only optimize the code but would also increase the accuracy when KNN is used. A couple of research papers have made use of PCA to implement feature selection, however, it might not be recommended in our research because, the data used is categorical and the selected features need to be displayed, which is extremely infeasible with the help of PCA.

Decision Tree used in [1] is a simple algorithm that might be used in our research, this machine learning model is generally used when a decision analysis needs to be made to identify a strategy to reach a goal. Our research has a similar goal i.e., we want to classify the data based on the features, and finally reach one category. Keeping these factors in mind, decision tree might seem like a fitting method, but due to the high number of features, i.e., 140+ features, the usage of decision tree becomes very inefficient, as the entropy of each feature needs to be calculated and based on this value, the tree subdivides. However, this might be used after applying feature selection, which would reduce the number of columns.

Zhexue Huang at.al., [7] in his research presented KModes, extended the K-Means algorithm to categorical domains and mixed numeric and categorical values. He presented the Kmodes algorithm which uses simple matching dissimilarity to deal with categorical data. With the help of this method, we will not have to encode our data to apply any clustering algorithm.

Devie Rosa Anamisa at.al.,[8] in her work proposed Naïve Bayes and Chi-square method to detect Acute Respiratory Infections in toddlers. She used Chi-square to reduce the number of features and include only necessary columns in the model. Naïve Bayes was incorporated to perform probability-based identification based on the application of the Bayes theorem with the assumption of strong independence.

M. Panda at.al., [8], [3], and [11] have made use of multinomial naive Bayes to build a network intrusion detection system. He used this method because the dataset they used was in categorical form. To apply a classification algorithm on a dataset that is in categorical form, the traditional algorithms might fail as they might be expecting a float type value, but they will receive a string type value. To avoid this error, multinomial naive Bayes is a very effective approach which also has a very low false positives rate and takes less time when compared to other algorithms.

Neha Sharma at.al.,[11] used K Modes algorithm for partition clustering. It takes randomly chosen initial cluster centres as seeds. This technique is widely applied to categorical data. The other algorithms select attributes based on frequency due to which they do not provide good results. K-modes selects algorithms based on information gain which helps in providing a better result and higher accuracy.

C. WORKFLOW DIAGRAM

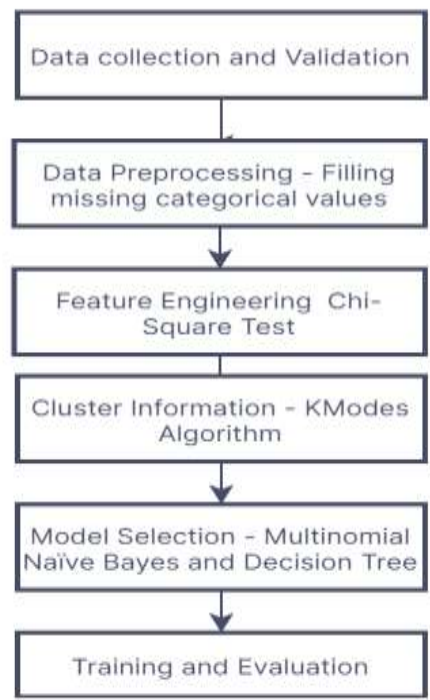

*Fig. 1 Workflow diagram of the entire model*

## II. METHODS

### A. DATA COLLECTION AND VALIDATION

Many considerations have been put forth to determine which characteristics of an individual are crucial in determining the 'Prakriti' of an individual. After studying enormous chunks of body features that impact the type of 'Prakriti' and scanning the various types of questionnaires published online by many websites, all important features are aggregated. Each body organ is comprehensively categorized further so that no organ cannot add any biases to our dataset. Evaluating these aspects, we collected our dataset from Mr. Arif Ahmed from his GitHub link given: https://github.com/skarifahmed/DeepAyurveda

We approached a few ayurvedic experts Dr. Milind Bidve (Pune, Maharashtra) to carry out initial data validity like the number of features, and sub-categorization of each feature. The initial three types of doshas were determined by the experts. Their approval from them triggered further steps in the construction of the model.

### B. DATA PRE-PROCESSING

The dataset needs to be cleaned to carry out assumptions on how each feature is being mapped to its corresponding 'Prakriti'. This includes the inspection of missing values in the dataset. There are many techniques to fill the missing values of categorical variables. Here, we use forward-filling technique of the Pandas library. This method is efficient as the dataset consists of categorical values.

### C. FEATURE ENGINEERING

Feature engineering helps to filter out some important features. The technique used is the Chi-square test. This test aims at gaining information about every feature type and how much they contribute to predicting the type of doshas because we are dealing with categorical data. We also use SelectKBest, which is a feature engineering technique to retrieve a ranked list of the most important features to the least important features. It facilitates us by eliminating redundant as well as irrelevant features.

*SelectKBest and Chi-Square Test*

SelectKBest is a feature selection technique that identifies the most important features from a dataset by evaluating the statistical relationship between each feature and target variable using a scoring metric. The higher the score, the more important that feature is. We can make use of different scoring metrics like Chi-square for categorical data and ANOVA for numerical data. Our dataset is in categorical form. Hence, we make use of the Chi-square test to select the 'K' best features.

Chi-Square test is generally used in cases where there is a significant association between the categorical variables used in the dataset. It considers the correlation of two variables as shown in (1). In our work, we intend to use the chi-square test during feature engineering as it will help filter the number of required attributes for building our model. The results given by this test will give the highest dependent features.

$$\chi^2 = \sum \frac{(O-E)^2}{E} \qquad (1)$$

$\chi^2$ = Chi Square Obtained
$\sum$ = Summation
O = Observed Score
E = Expected Score

The implementation of the chi-square test can be done by using the chi library from the sklearn feature selection. The test gives us two values. The values are the F-score and the P-values. The F-score is used to evaluate the performance of this model, whereas the P-values are used to accept or reject the null hypothesis assumed. Higher F-score values suggest higher importance of the feature. P-values have inverse relation in establishing the importance. Lesser The P-values, greater is the importance of that feature.

### D. CLUSTER INFORMATION

One of the fundamental aspects of this study is the expansion of the existing three-segment classification of 'Prakriti' types into seven segments. The seven segments correspond mainly to each existing 'Prakriti' type and then pairing them. Combining all such pairs we have 'Pure Kapha', 'Pure Vata', 'Pure Pita', 'Kapha-Vata', 'Vata-Pita' and 'Pita-Kapha'. This cluster formation is essential to devise Ayurvedic medicines.

*K-Modes*

The dataset that we have consists of copious categorical features. To group these types into clusters, an algorithm is devised which takes into consideration the categorical data. The clustering of data is done with the help of K-modes clustering. The reason for choosing such an algorithm is because it works like the popular KMeans clustering algorithm but for categorical data, it considers the mode of each attribute. The iterative process accounts for dissimilarities in the categorical data. It assigns a centroid as the most frequent value (mode). The process lasts till we find perfect centroid values. To use this strategy in our work, we provide the 7 clusters as output for the model. Each iteration calculates the dissimilarities (cost) and the number of moves. The iterations run up to a default value of 100.

### E. MODEL SELECTION

The categorization of the data where we had only three 'Prakriti types' – 'Vata', 'Pita', and 'Kapha' is enhanced into seven minor clusters, to provide better clarity about the 'Prakriti' of a person. The clusters that have been derived from the parent 'Prakriti' types need to be evaluated for correctness. Here we have tried the Multinomial Naïve Bayes (MNB) classifier and Decision tree classifier.

### F. TRAINING AND EVALUATION

To prepare the data for our model, we need to segregate it. This is done to avoid overfitting during model evaluation.

We propose the training and testing of data by using the inbuilt Scikit-learn library train_test_split. Initially, we split the data such that we divide the data into two chunks. These chunks have 80% data for training and 20% data for testing. The training data consists of the tuples we grouped having mixed 'Prakriti' types. This data is to be fitted into the model. The testing data on the other hand has the same set. It would be used to evaluate the accuracy of the model by calculating the accuracy, precision, F-score, and recall.

*MNB Classifier*

The Multinomial Naïve Bayes classifier is designed to classify categorical values. It employs the Bayes theorem to predict the class for a new instance. It accepts training data and fits it into a model. It works as shown in (2).

$$P(x) = \frac{P(c)\,P(c)}{P(x)} \quad (2)$$

$P(x)$ = Posterior Probability
$P(c)$ = Likelihood
$P(c)$ = Class Prior Probability
$P(x)$ = Predictor Prior Probability

Here, the reason for choosing the model is, it is most suitable for handling categorical data and evaluating the word frequency as we have in our dataset.

*Decision Tree*

A supervised learning model that represents a sequence of decisions and their possible consequences. It splits the data based on feature values, starting from the root node, and branching out to internal nodes. At each internal node, a decision is made using a set of given features, this determines the path to be followed. This process continues recursively until reaching leaf nodes, which represent the final classification or regression outcome. The reason for choosing this algorithm is, to deal with the high cardinality of the dataset. Another reason to use this is the ability to post-prune so that the full-grown tree having some undermining branches could be retracted to improve overall performance.

### III. PERFORMANCE EVALUATION

*INITIAL ANALYSIS*

To evaluate our model, we consider two factors. Size of the testing datasets and the number of features considered. We check these factors to understand how our model performs with different parameters and find out the best working model along with parameters to analyse the performance of different machine learning techniques. The models used are MNB classifier and Decision tree. We take the reduced features from the Chi-square feature reduction. We analyse and study the behaviour for two cases – having test size as 10% of the dataset and then 20% of the dataset.

*Accuracy*

Accuracy is an essential metric as it calculates the overall correctness of the predicted values gained by our precision models. It represents a ratio of correctly classified instances to the total number of instances in the dataset as in (3).

$$\frac{(True\ positive + True\ Negative)}{True\ Positive + True\ Negative + False\ Positive + False\ Negative} \quad (3)$$

*Precision*

Precision gives the proportion of correctly predicted values out of all the predicted trues by our model. It is very essential to understand the use of precision in evaluating our model as it focuses on the correctness of positive predictions. It is also an essential metric when the cost of false positives is high. It is calculated as the ratio of true positives to the sum of true positives and false positives as shown in (4).

$$Precision = \frac{True\ positives}{True\ positives + False\ positives} \quad (4)$$

*Recall*

Recall focuses on the model's ability to detect the positive instances in the dataset. It is the ratio of the correct positive instances to the total number of positive instances. It is essential to use precision as it is independent of any negative instance classification in our dataset. Formula is given in (5).

$$Recall = \frac{True\ positives}{True\ positives + False\ Negatives} \quad (5)$$

*F-score*

It is the harmonic mean of precision and recall. It is calculated as 2 times the product of precision and recall, divided by the sum of precision and recall. It gives us a trade-off between both parameters. A high F-score suggests a good and balanced model. Formula is given in (6).

$$F1\ Score = \frac{2*(Precision*Recall)}{Precision + Recall} \quad (6)$$

### B. RESULTS

In our work we proposed two models for MNB classifier and decision tree. These models are fitted with categorical data and are used to retrieve the predicted values for the testing data. The predicted values obtained will be used for evaluation. The evaluation metrics considered by us are accuracy, precision, f score, and recall. These metrics give us a deep insight into the working of our models. The test size is also taken into consideration while building the evaluation metrics. This is primarily done because test size influences the ability to generalize the model's performance. Furthermore, any imbalances in the dataset are handled neatly. Adjusting test size at regular intervals helps us to negate any imbalances that may occur during model training and testing.

*MNB Classifier*

We evaluate the performance of both models by a comparative approach. The results obtained in the MNB classifier for test-size as 0.1 and 0.2 are considered. The number of features in both test-size gradually increased from

20 to 100. The accuracy, precision, F-score, and Recall are calculated for each. A tabulated approach to study this is presented in TABLE I.

TABLE I
Performance of MNB classifier

| Test size | No. of features | Accuracy | Precision | F-Score | Recall |
|---|---|---|---|---|---|
| 0.1 | 20 | 0.93 | 0.73 | 0.90 | 0.93 |
|  | 40 | 0.73 | 0.63 | 0.75 | 0.73 |
|  | 60 | 0.53 | 0.38 | 0.51 | 0.53 |
|  | 80 | 0.86 | 0.91 | 0.86 | 0.86 |
|  | 100 | 0.80 | 0.81 | 0.79 | 0.80 |
| 0.2 | 20 | 0.90 | 0.80 | 0.90 | 0.90 |
|  | 40 | 0.83 | 0.78 | 0.84 | 0.83 |
|  | 60 | 0.70 | 0.61 | 0.66 | 0.70 |
|  | 80 | 0.83 | 0.91 | 0.80 | 0.83 |
|  | 100 | 0.86 | 0.87 | 0.86 | 0.86 |

The two testing sizes are varied according to the number of features and the changes in metrics are observed. When features are 20, we see that the accuracy is highest, i.e., 0.93, and it drops a little when we increase the number of features and finally settles to 0.80. The F-score which is responsible for a trade-off between precision and recall also decreases with the number of features. The highest value for F-score is 0.90 and the least is 0.79. Similar observations can be made for the test size 0.2. However, it is germane to discuss that, precision in this case is higher than the one in test size 0.1 because false positives significantly decrease in this case as the dataset increases.

*Decision Tree Classifier*

The decision tree classifier is used to build models to understand a comparative difference between the former MNB classifier and the Decision tree classifier. Similar results can be obtained when we iteratively check for features from 20 to 100. The accuracy, precision, F-score, and recall are calculated in each case. The trend is similar, where, on increasing the number of most important features, we see a decreasing trend in these metrics.

For the first case we had a testing set of 0.1. The accuracy was highest for 20 features i.e., 0.73, and least for 100 i.e., 0.46. The precision and recall decrease from 0.75 to 0.38 and 0.73 to 0.46 respectively. F-score helps us understand the trade-off between these two. This is clearly shown in TABLE II.

TABLE II
Performance of Decision tree classifier

| Test size | No. of features | Accuracy | Precision | F-Score | Recall |
|---|---|---|---|---|---|
| 0.1 | 20 | 0.73 | 0.75 | 0.75 | 0.73 |
|  | 40 | 0.60 | 0.58 | 0.62 | 0.60 |
|  | 60 | 0.40 | 0.33 | 0.40 | 0.40 |
|  | 80 | 0.66 | 0.66 | 0.64 | 0.66 |
|  | 100 | 0.46 | 0.38 | 0.51 | 0.46 |
| 0.2 | 20 | 0.76 | 0.66 | 0.76 | 0.76 |
|  | 40 | 0.70 | 0.79 | 0.71 | 0.70 |
|  | 60 | 0.50 | 0.43 | 0.48 | 0.50 |
|  | 80 | 0.53 | 0.40 | 0.47 | 0.53 |
|  | 100 | 0.66 | 0.50 | 0.68 | 0.66 |

For the second case, when the test size is 0.2, the metrics performance is better. As we increased number of features from 20 to 100, the accuracy decreases from 0.76 to 0.66. Due to the erratic nature of both precision and recall, we evaluate the F-score. The F-score like other metrics is found to decrease a little and then increase for 100 features.

The comparative analysis of both models shows that for many cases, the metrics performed better for the MNB classifier than for the decision tree classifier. This is the perfect result that helps us understand a better way to develop clustering models.

*C. Discussion*

The goal of our research work is to effectively classify a person into the seven doshas that are 'Vata', 'Pita', 'Kapha', 'Vata-Kapha', 'Vata-Pita', 'Pita-Kapha', 'Vata-Pita-Kapha' based on selected features of the body. The method applied to achieve this has been articulately described in our work.
In the results, we have effectively tested two algorithms namely Multinomial Naïve Bayes and Decision tree, against four major criteria namely accuracy, precision, F-Score, and recall. This evaluation scheme has provided us with the differences between both the algorithms and the reasons why we would choose the MNB algorithm over Decision tree.
Our findings are different from existing work in this field as we have not confined ourselves to 3 'Doshas', instead, we have classified a person into 7 doshas based on their body features. In the past, researchers have made use of SVM [1], [3] and KNN [1], [2] and [3] as the dataset they used was encoded. Algorithms proposed in [6] and [9] made use of K-modes as their data was categorical but did not use any specific feature selection technique which would boost their accuracy. The dataset that we have chosen had 147 features. Out of these features, as not all played a significant role in determining the 'Prakriti' of a person, we made use of Chi-Square test to make sure that the features we are selecting entirely make an impact on the 'Dosha' of a person. The number of features we are taking into consideration also creates a significant impact on the accuracy of the model.

These findings can be helpful for people who want to assess their bodies keeping the methods and benefits of Ayurveda as their main goal. This research would make it easier for Ayurvedic practitioners to know the 'Dosha' of a person and then conclude on the kind of lifestyle one should follow. The process becomes less time-consuming.

*Multinomial Naïve Bayes*

When the test size was 0.1 and number of features was varied, the bar plot was as seen in Fig. 2. With 20 features, the accuracy was 0.93, precision was 0.73, F-score was 0.90 and recall was 0.93. The accuracy is high because number of features is less. As a direct consequence, only the most important features were considered, which made the classification simple. With 40 features, accuracy was 0.73, precision was 0.65, F-score was 0.75 and recall was 0.73. This sudden decrease in the above metrics was due to the increase in the number of features. These features do not impact the 'Prakriti' type as the former 20 features. When we increase the number of features to 60. Accuracy was 0.53, precision was 0.38, F-score was 0.51 and recall was 0.53. This was due to the inundated data. Further, the number of features was increased to 80. This gave us an accuracy of 0.86, precision of 0.91, F-score of 0.86, and a recall of 0.86. This was the best parameter as it yielded the maximum accuracy and precision. This was because majority of the 80 features selected were significant. Lastly, with 100 features, accuracy was 0.80, precision was 0.81, F-score was 0.79 and recall was 0.80. This was lesser when compared to 80 features. This is because of the inclusion of features that did not profoundly impact the 'Prakriti'.

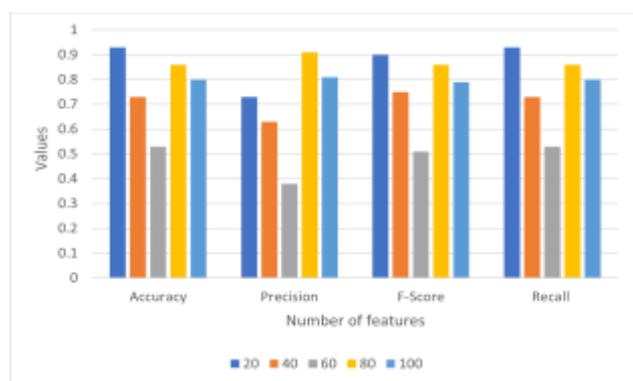

*Fig. 2* Bar graph of Number of features Vs metrics test size at 10%

Test size fixed to 0.20, graph is shown in Fig. 3. With 20 features, the accuracy was 0.90, precision was 0.80, F-score was 0.90 and recall was 0.90. The accuracy was lesser than the value when test size was 0.10. This is because here the rows that are selected are random. So, when the test size increases, accuracy is affected. With 40 features accuracy was 0.83, precision was 0.78, F-score was 0.84 and recall was 0.83. This decrease in accuracy and precision was due to increment in the features taken. There might be some features that do not bother the 'Prakriti'. With 60 features we got the least accuracy of 0.70, precision of 0.61, F-score of 0.66, and recall of 0.70. This is due to the inclusion of unnecessary features in the model. Next, with 80 features, accuracy of 0.83, precision 0.91, F-score of 0.80 and recall of 0.83. These numbers depict that the number of true positives out of all the trues was very high here. With 100 features, the accuracy was 0.86, precision was 0.87, F-score was 0.86 and the recall was 0.86. These are less than the metrics we received when number of features was 80, due to irrelevant features.

From the above metrics, we conclude that Multinomial naïve Bayes is better than decision tree because our dataset has irrelevant information and inconsistencies. This makes the dataset very noisy. So, it is highly recommended to use multinomial naïve Bayes over decision tree. Additionally, our dataset comprises 147 features. This makes the dataset high dimensional because of which, naïve Bayes outperforms decision tree.

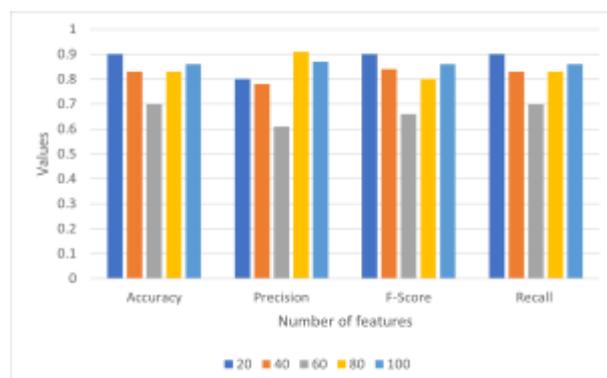

*Fig. 3* Bar graph of Number of features Vs metrics test size at 20%

## CONCLUSION

The purpose of our research has been aptly described in our work. We bridged the gap between establishing a closer relationship between human behaviour and 'Prakriti'. The works in the past have considered only three classes but we made it feasible enough to substantially increase our outlook from three 'Doshas' to seven 'Doshas'. After combining various parameters, we can find a balance between the parameters which give the best possible results. To better understand the implications of our study, future research could implore gaining knowledge about a better as well as more efficient way to group these 'Dosha' types. Our work can thus be used to understand the 'Prakriti' type and give the best possible recommendations based on them to any person. This would help revive and improve the ancient systems of Ayurveda practiced in India and bring a boom to the advancements of medicine.